\documentclass{opt2021} 

\title[Structured Low-Rank Tensor Learning]{Structured Low-Rank Tensor Learning}

\usepackage{enumerate}

\optauthor{%
\Name{Jayadev Naram} \Email{jayadev.naram@research.iiit.ac.in}\\
\Name{Tanmay Kumar Sinha} \Email{tanmay.kumar@research.iiit.ac.in}\\
\Name{Pawan Kumar} \Email{pawan.kumar@iiit.ac.in}\\
\addr CSTAR, IIIT Hyderabad}

\begin{document}

\maketitle

\newcommand{\Norm}[1]{\left\lVert#1\right\rVert}

\newcommand{\R}{\mathbb{R}}
\newcommand{\W}{\mathcal{W}}
\newcommand{\Y}{\mathcal{Y}}
\newcommand{\Z}{\mathcal{Z}}
\newcommand{\s}{\mathcal{S}}
\newcommand{\Rtensor}{\R^{n_1\times \cdots \times n_K}}
\newcommand{\Rptensor}{\R_+^{n_1\times \cdots \times n_K}}
\newcommand{\A}{\mathcal{A}}
\newcommand{\M}{\mathcal{M}}
\newcommand{\N}{\mathcal{N}}
\newcommand{\h}{\mathcal{H}}
\newcommand{\T}{\mathcal{T}}
\newcommand{\Prob}{\mathbb{P}}
\newcommand{\Dist}{\mathcal{D}}
\newcommand{\perpProj}{P^\perp}
\newcommand{\bb}{\mathbb{B}}
\newcommand{\Sprod}{\mathbb{S}_{xy}}
\newcommand{\highlight}[1]{\textsl{\textbf{#1}}}
\newcommand{\mapping}[3]{#1:#2\rightarrow #3}
\newcommand{\doubt}{\highlight{[??]}}
\newcommand{\bigvert}[2]{\left.#1\right|_{#2}}
\newcommand{\sdnn}[1]{${#1}$}
\newcommand{\bsdnn}[1]{$\boldsymbol{#1}$}
\newcommand{\ifthen}[2]{\textbf{(#1)}\boldsymbol{\implies}\textbf{(#2)}}
\newcommand{\bsdn}[1]{\boldsymbol{#1}}
\newcommand{\forward}{$(\implies)$}
\newcommand{\converse}{$(\impliedby)$}
\newcommand{\Lt}[1]{\underset{#1\rightarrow 0}{Lt}}
\newcommand{\norm}[1]{\|#1\|}
\newcommand{\dparder}[2]{\dfrac{\partial #1}{\partial x_{#2}}}
\newcommand{\fparder}[2]{\frac{\partial #1}{\partial x_{#2}}}
\newcommand{\parder}[2]{\partial #1/\partial x_{#2}}
\newcommand{\parop}[1]{\dfrac{\partial}{\partial x_{#1}}}
\newcommand{\innerproduct}[2]{\langle #1, #2 \rangle}
\newcommand{\genst}{St_B(n,p)}
\newcommand{\igenst}[1]{St_{B_{#1}}(n_{#1},p)}
\newcommand{\realmat}[2]{\R^{#1\times #2}}
\newcommand{\Skew}{\mathcal{S}_{skew}(p)}
\newcommand{\Sym}{\mathcal{S}_{sym}(p)}
\newcommand{\XperpB}{X_{B^\perp}}
\newcommand{\polarRetr}{R^{polar}_X}
\newcommand{\qrRetr}{R^{QR}_X}
\newcommand{\vectransport}{\mathcal{T}}
\newcommand{\grad}{\text{grad}\,}
\newcommand{\hess}{\text{Hess}\,}
\newcommand{\unfold}[1]{\textit{unfold}_{#1}}
\newcommand{\fold}[1]{\textit{fold}_{#1}}

\setlength{\abovedisplayskip}{3pt}
\setlength{\belowdisplayskip}{3pt}

\begin{abstract}%
We consider the problem of learning low-rank tensors from partial observations with structural constraints, and propose a novel factorization of such tensors, which leads to a simpler optimization problem. The resulting problem is an optimization problem on manifolds. We develop first-order and second-order Riemannian optimization algorithms to solve it. The duality gap for the resulting problem is derived, and we experimentally verify the correctness of the proposed algorithm. We demonstrate the algorithm on nonnegative constraints and Hankel constraints.
\end{abstract}



\section{Introduction}

With the rise in the availability of multidimensional data such as colour images, video sequences, and hyperspectral images, tensor-based techniques have started gaining attention as traditional matrix-based methods cannot exploit the underlying structure present in higher-dimensional data. Many recent applications of tensor reconstruction techniques also enforce structural constraints such as nonnegativity (\cite{ncpc}, \cite{nonneg_ieee_access}) or a Hankel structure (\cite{missing_slice}, \cite{exponential_signals}).

We propose a framework for dealing with tensor completion problems with general structural constraints. In particular, we consider the structured low-rank tensor learning problem of the following form:
\begin{equation}
\begin{aligned}
\underset{\W\in\Rtensor}{\text{min}}& &&C\|\W_\Omega-\Y_\Omega\|^2 + R(\W)\\
\text{subject to}&\hspace{0.5cm} &&A(\W) \ge 0, \label{eqn:primal_problem}
\end{aligned}
\end{equation}
where $\Y_\Omega\in \Rtensor$ is a partially observed tensor for indices given in the set $\Omega$, $(\W_\Omega)_{i_1,\ldots,i_K} = w_{i_1,\ldots,i_K}$ for $({i_1,\ldots,i_K})\in \Omega$, $C>0$ denotes the cost parameter, $\mapping{R}{\Rtensor}{\R}$ is a regularizer that induces low-rank constraint on the tensor, and $\mapping{A}{\Rtensor}{\R^n}$ is a linear map that induces structural constraint on the tensor.

Following \cite{dual}, we learn the tensor as a sum of $K$ tensors, $\W = \W^{(1)} + \cdots + \W^{(K)}$, and use the following low-rank regularizer:
\begin{equation}
R(\W) = \sum_{k=1}^K \dfrac{1}{\lambda_k}\Norm{W_k^{(k)}}^2_*.
\end{equation}
The main contributions of the paper are given below.
\begin{itemize}
    \item We propose a novel factorization for modeling structured low-rank tensors through a partial dual problem of \eqref{eqn:primal_problem}.
    \item We develop first-order and second-order Riemannian optimization algorithms that exploit proposed factorization's inherent geometric structure.
    \item We compute the expression for the duality gap and verify the correctness of the proposed algorithm through experiments.
    \item We apply the proposed algorithm to the nonnegative constraint and the Hankel constraint.
\end{itemize}

\section{Notation}

We follow \cite{kolda_review} for our tensor notation. We present a few important notions here. 
Tensors are denoted by uppercase calligraphic letters, e.g., $\W$. Matrices are denoted by uppercase letters, e.g., $X$. For a square matrix $X\in\R^{n\times n}$, we denote its trace by $tr(X)$. $\R_+$ denotes the interval $[0,\infty)$.
The inner product of two tensors is defined as follows: 
\begin{equation*}
    \innerproduct{\W}{\mathcal{U}} = \sum_{i_1=1}^{n_1}\sum_{i_2=1}^{n_2}\cdots \sum_{i_K=1}^{n_K} w_{i_1,\ldots,i_K}u_{i_1,\ldots,i_K}.
\end{equation*}
A mode-$k$ fiber of a tensor $\W\in\Rtensor$, denoted by $w_{i_1,\ldots,i_{k-1},:,i_{k+1},\ldots,i_K}$, is a vector obtained by fixing all but $k$-th index of $W$. The mode-$k$ unfolding of a tensor $\W\in\Rtensor$ is a matrix $unfold_k(\W) = W_k\in \R^{n_k\times n_1\cdots n_{k-1}n_{k+1}\cdots n_K}$ formed by arranging the mode-$k$ fibers to be the columns of the resulting matrix. Similarly, we can define the $k$-mode folding operation ($fold_k$) as the inverse of the unfolding operation - it converts a given matrix to a tensor of a suitable order. 
The $k$-mode product of a tensor $\W \in \Rtensor$ with a matrix $U \in \R^{m \times n_k}$ is defined as follows:
\begin{equation*}
    (\W\times_k U)_{i_1,\ldots,i_{k-1},j,i_{k+1},\ldots,i_K} = \sum_{i_k}^{n_k} w_{i_1,\ldots,i_K}u_{j,i_k}, \;\text{i.e.},\;\mathcal{X} = \W \times_k U \iff X_k = U W_k.
\end{equation*}

\section{Related Work}

There are a number of well-known methods for tensor completion using different approaches for enforcing the low-rank constraint. Many methods use a generalization of the matrix trace norm regularizer to the tensor case. Two well-known regularizers are the overlapped trace norm regularizer (\cite{oldest_lrtc}, \cite{scalable_tensor_learning} and \cite{spectral_regularization}) and latent trace norm regularizer (\cite{ffw_lrtc}). The overlapped trace norm regularizer  uses the regularizer $R(\W) = \sum_{k=1}^{K} \|W_k\|_*$. In the latent trace norm regularizer model, the tensor is modelled as the sum of $K$-tensors, $\W = \sum_{k=1}^K \W^{(k)}\in\Rtensor$, and the regularizer is defined as
\begin{equation*}
    R(\W) = \inf_{\substack{\W^{(k)},\\ k \in \{1,\ldots,K\}}} \sum_{k=1}^{K} \Norm{W^{(k)}_k}_*.
\end{equation*}

Other methods for tensor completion use tensor decompositions like the Tucker decomposition and CP decomposition to learn the tensors. \cite{geomCG} and \cite{RPrecon} formulate the tensor completion problem as an optimization problem on the Riemannian manifold of fixed multi-linear rank tensors. 

A general formulation for low-rank matrix completion problems with structural constraints was developed in \cite{structured_matrix_completion}. It provided a unified framework for dealing with general linear inequality and equality constraints. 

Recently, tensor completion problems with structural constraints have started attracting attention. \cite{ncpc} and \cite{nonneg_ieee_access} develop nonnegative tensor completion algorithms for image and video reconstruction tasks. \cite{nonneg_wacv} falls under the current framework which considers the special case of nonnegative constraints in detail with vast experimentation. \cite{missing_slice} models image completion with missing slices as a higher-order Hankel tensor completion problem. \cite{traffic_speed} uses low-rank Hankel tensor model for estimating traffic states from partial observations. 

\section{Dual Framework}

Following \cite{dual}, we construct a partial dual problem to the primal problem \eqref{eqn:primal_problem}, incorporating the structural constraint into the formulation. We do this using the approach outlined in \cite{structured_matrix_completion}. We generalize Theorem 1 of \cite{dual} to the case of tensors with structural constraints. The following lemma \cite{lemma_proof} is used in the development of the dual formulation.

\begin{lemma}\label{lemma:nuclear_norm}
For a matrix $X \in \R^{d \times T}$, the nuclear norm of $X$ satisfies the following relation:
\begin{equation*}
    \|X\|_*^2 = \min_{\Theta \in P^d, \, \text{range}(X) \subseteq \text{range}(\Theta)} \innerproduct{\Theta^{\dagger}X}{X},
\end{equation*} where $P^d = \{ S \in \R^{d \times d}:\, S \succeq 0, \text{tr}(S) = 1\} $, $\text{range}(\Theta) = \{\Theta z:\, z \in \R^d\}$, $\Theta^\dagger$ denotes the pseudo-inverse of $\Theta$. For a given $X$, the minimizer is $\bar{\Theta} = \sqrt{XX^T} / \text{tr}(\sqrt{XX^T}) $.
\end{lemma}

Using the above lemma, we can write \eqref{eqn:primal_problem} as 
\begin{equation}
\begin{aligned}
\underset{\substack{\Theta_k \in P^{n_k},\\ k \in \{1,\ldots,K\}}}{\text{min}}\;\underset{\substack{\W^{(k)},\\ k \in \{1,\ldots,K\}}}{\text{min}}& &&C\Norm{\bigg(\sum_{k=1}^K \W^{(k)}\bigg)_\Omega-\Y_\Omega}^2 + \sum_{k=1}^K \frac{1}{2\lambda_k} \innerproduct{\Theta_k^\dagger W^{(k)}_k}{W^{(k)}_k}\\
\text{subject to}&\hspace{0.5cm} &&A(\W) \ge 0. \label{eqn:primal_rewritten}
\end{aligned}
\end{equation}

\begin{theorem}\label{thm:dual_form}
The following minimax problem is equivalent to the problem \eqref{eqn:primal_rewritten}:
\begin{alignat}{3}\label{eqn:minimax_problem}
\underset{\substack{\Theta_k\in P^{n_k},\\ k \in \{1,\dots,K\}}}{\min}& \underset{\Z\in\mathcal{C}, \, s\in\R^n}{\max}&\; && \innerproduct{\Z}{\Y_\Omega} - \dfrac{1}{4C}\|\Z\|^2 -\sum_{k=1}^K \frac{\lambda_k}{2} \innerproduct{Z_k + (A^*(s))_k}{\Theta_k[Z_k + (A^*(s))_k]},
\end{alignat}
where $\mathcal{C} = \{\Z\in\Rtensor:\,\Z = \Z_\Omega\}$.
%
The optimal solution $\bar{\W}$ of \eqref{eqn:primal_rewritten} is related to the optimal solution $\{\bar{\Theta}_1,\ldots,\bar{\Theta}_K,\bar{\Z},\bar{s}\}$ of \eqref{eqn:minimax_problem} by 
\begin{equation}\label{eqn:primal_dual_relation}
\bar{\W} = \sum_{k=1}^K \lambda_k(\bar{\Z}+A^*(\bar{s}))\times_k\bar{\Theta}_k.
\end{equation}
\end{theorem}
For proof, see Appendix \ref{sec:dual_proof}.


From \eqref{eqn:primal_dual_relation} it can be seen that the structured and low-rank constraints on $\W$ can be decomposed into structured constraints on $s$ and low-rank constraints on $\Theta_k,$ which leads to a simpler optimization method.

\section{Proposed Algorithm}

Due to the low-rank constraint on $\W$, each $\Theta_k$ has a low rank. Therefore, a fixed-rank parameterized problem can be given by writing $\Theta_k = U_k U_k^T$. The problem \eqref{eqn:minimax_problem} can then be written as follows.
\begin{equation}\label{eqn:outer_manifold_problem}
\underset{U\in S^{n_1}_{r_1}\times\cdots\times S^{n_K}_{r_K}}{\min}\;g(U),
\end{equation}
where $U = (U_1,\ldots,U_K)$, $S^{n}_{r} = \{U\in\R^{n\times r}:\,\|U\|_F = 1\},$ and 
\begin{equation}\label{eqn:inner_strong_convex_problem}
g(U) = \underset{\Z\in\mathcal{C},s\in\R^n}{\max}\; \innerproduct{\Z}{\Y_\Omega} - \dfrac{1}{4C}\|\Z\|^2 -\sum_{k=1}^K \frac{\lambda_k}{2} \Norm{U_k^T(Z_k + (A^*(s))_k)}^2.
\end{equation}
The optimization problem in \eqref{eqn:inner_strong_convex_problem} is strongly convex for a given $U$, while problem \eqref{eqn:outer_manifold_problem} is a non-convex problem in $U$.

The set $S^{n_1}_{r_1}\times\cdots\times S^{n_K}_{r_K}$ is a Riemannian manifold (\cite{absil_spectrahedron}, \cite{dual}), and thus problem \eqref{eqn:outer_manifold_problem} is an optimization problem on a manifold. We solve it using either Riemannian conjugate gradient or Riemannian Trust-Region algorithm, depending on the structural constraint. The proposed algorithm is shown in Algo.~\ref{algo:proposal}. For more details on optimization on general manifolds, we refer the reader to \cite{boumal_book} and \cite{absil_book}. 

The Riemannian optimization algorithm in Algo.~\ref{algo:proposal} requires computing the Euclidean gradient and its directional derivative, which are given in the following lemma.

\begin{lemma}\label{lemma:grad_hessian}
Let $\{\hat{\Z},\hat{s}\}$ be the maximizer of the convex problem \eqref{eqn:inner_strong_convex_problem} at $U$. Then, the Euclidean gradient $\nabla g(U)$ is given by
\begin{equation*}
    \nabla g(U) = - (\lambda_1 P_1, \ldots, \lambda_K P_K),
\end{equation*}
where $P_k = (\hat{Z}_k + (A^*(\hat{s}))_k)(\hat{Z}_k+(A^*(\hat{s}))_k)^T U_k$. Let $V\in \R^{n_1\times r_1}\times\cdots\times \R^{n_K\times r_K}$ and $\Dot{Z}_k, \Dot{s}$ denote the directional derivatives of $Z_k$ and $s$ along $V$ respectively. Then, the directional derivative of $\nabla g$ at $U$ along $V$ is
\begin{equation*}
    D\nabla g(U)[V] = - (\lambda_1 Q_1, \ldots, \lambda_K Q_K),
\end{equation*}
where $Q_k = (\hat{Z}_k + (A^*(\hat{s}))_k)(\hat{Z}_k+(A^*(\hat{s}))_k)^T V_k + 2 \, sym((\Dot{Z}_k + (A^*(\Dot{s}))_k)(\hat{Z}_k+(A^*(\hat{s}))_k)^T))U_k$ and $sym(X) = (X+X^T)/2$.
\end{lemma}

\begin{remark}\label{remark:Zsdot}
It can be seen that computing $D\nabla g(U)[V]$ requires the terms $\Dot{\Z}$ and $\Dot{s}$. These terms can be computed by applying directional derivative along $V$ on the first-order optimality conditions of the problem \eqref{eqn:inner_strong_convex_problem} at $\{\hat{\Z},\hat{s}\}$.
\end{remark}

\begin{algorithm2e}\label{algo:proposal}
 \caption{Proposed Algorithm for Structured Low-Rank Tensor Completion}
 \SetAlgoLined
  \KwData{$\Y_\Omega$, rank=$(r_1,\ldots,r_K)$, $\varepsilon$, $(\lambda_1,\ldots,\lambda_K)$}
  \KwResult{$\hat{\W} = \sum_{k=1}^{K} \lambda_k(\hat{\Z}+A^*(\hat{s})) \times_k (U_k U_k^T)$}
  	\For{$t = 1,2,\cdots$}{
	 Check Termination: if $\|\nabla g(U^{(t)})\| \le \varepsilon$ then break\;
    Solve for $\hat{\Z}^{(t)}$ and $\hat{s}^{(t)}$ in \eqref{eqn:inner_strong_convex_problem}\;
	Compute cost $g(U^{(t)})$, gradient $\nabla g(U^{(t)})$ and the directional derivative of $\nabla g(U^{(t)})$\;
	Update $U$: $U^{(t+1)}$ = {\tt RiemannianCG-update}$(U^{(t)})$ or\\ \hspace{1.7cm}$U^{(t+1)}$ = {\tt RiemannianTR-update}$(U^{(t)})$\;
    }
\end{algorithm2e}

We have the following result regarding the optimality of the proposed algorithm. It is a generalization of Theorem 3 in \cite{structured_matrix_completion}.

\begin{theorem}\label{theorem:duality_gap}
Let $\hat{U} = \left(U_1, \ldots, U_K \right)$ be a feasible solution of problem \eqref{eqn:outer_manifold_problem} and $\left\{ \hat{\Z},\hat{s}\right\}$ be the maximizer of the convex problem \eqref{eqn:inner_strong_convex_problem} at $U = \hat{U}$. Let $\hat{\A} = A^{*}(\hat{s})$, and $\sigma_k$ be the maximum singular value of $\hat{Z}_k + \hat{A}_k$. Let $\hat{\Theta} = (\hat{\Theta}_1,\ldots,\hat{\Theta}_K)$, where $\hat{\Theta}_k = \hat{U}_k \hat{U}_k^T$. Then,
$\left\{\hat{\Theta},\hat{\Z},\hat{s}\right\}$ is a candidate solution for the partial dual problem \eqref{eqn:minimax_problem} and we have the following expression for the duality gap $\Delta$:
\begin{equation}
\Delta = \sum_{k=1}^{K} \frac{\lambda_k}{2} \left( \sigma_k^2 -  \|\hat{U}_k^T(\hat{Z}_k+\hat{A}_k)\|^2 \right).\label{eqn:duality_gap}
\end{equation}
\end{theorem}
For proof see Appendix \ref{sec:duality_gap_proof}.

\section{Applications}

We consider several popular applications for our proposed method. See \cite{dual} for the case where there are no structure constraints.

\subsection{Nonnegative Tensor completion}

The nonnegative tensor completion problem is 
\begin{equation}
\begin{aligned}
\underset{\W\in\Rtensor}{\text{min}}& &&C\|\W_\Omega - \Y_\Omega\|^2 + \sum_{k=1}^K \dfrac{1}{\lambda_k}\Norm{W_k^{(k)}}^2_* \\
\text{subject to}&\hspace{0.5cm} &&\W \ge 0. \label{eqn:primal_problem_nn}
\end{aligned}
\end{equation}
The fixed-rank dual problem of \eqref{eqn:primal_problem_nn} is
\begin{equation}\label{eqn:outer_manifold_problem_nn}
\underset{U\in S^{n_1}_{r_1}\times\cdots\times S^{n_K}_{r_K}}{\min}\;g(U),
\end{equation}

$g(U)$ is given by

\begin{equation}\label{eqn:inner_strong_convex_problem_nn}
g(U) = \underset{\s\in\Rptensor}{\max}\bigg(\underset{\Z\in\mathcal{C}}{\max}\; \innerproduct{\Z}{\Y_\Omega} - \dfrac{1}{4C}\|\Z\|^2 -\sum_{k=1}^K \frac{\lambda_k}{2} \Norm{U_k^TZ_k + U_k^TS_k}^2\bigg).
\end{equation} where $U = (U_1,\ldots,U_K)$ and $\Rptensor$ is the set of all tensors of size $n_1 \times \cdots n_K$ with non-negative entries.

The problem \eqref{eqn:inner_strong_convex_problem_nn} can be solved alternatively for $\Z$ and $\s$. Then the resulting problem in $\Z$ is unconstrained least-squares problem, which can be solved using linear conjugate gradient algorithm (For various preconditioned CG approaches, see \cite{saad2003,benzi2002,das2020,das2021,katyan2020,mehta2020,kumar2014,kumar2016,kumar2013,kumar2013b,rampalli2018,kumar2010c,aggarwal2019,kumar2015b,kumar2010d,kumar2015e,niu2010,kumar2011c,kumar2014s}), and over $\s$ it is non-negative least-squares problem which can be solved using \cite{nnls_sra}. The cost $g(U)$ and the gradient $\nabla g(U)$ can be easily computed following Lemma \ref{lemma:grad_hessian}, hence, we employ Riemannian conjugate gradient to solve the outer optimization problem over $U$. The per-iteration complexity of the proposed algorithm is $
O\bigg(T|\Omega|\sum_{k=1}^K{r_k}+\sum_{k=1}^K n_k r_k^2+\sum_{k=1}^K r_k^3\bigg)$, where $T$ is the sum of iterations of CG and NNLS solver for solving $\Z$ and $\s$ respectively.

\subsection{Hankel Tensor Completion}

The primal problem is given by
\begin{alignat}{3}\label{eqn:hankel_primal}
\underset{\W\in\Rtensor}{\text{min}}& &&C\|\W_\Omega-\Y_\Omega\|^2 + \sum_{k=1}^K \dfrac{1}{\lambda_k}\Norm{\h_{2k}(\W^{(k)})}^2_*,
\end{alignat}
where $\h_k = \unfold{k}\circ \h$ and $\h$ is the $k$-th order Hankel transform such that for $\W\in \Rtensor$, $\h(\W)\in \R^{\tau_1\times n_1-\tau_1+1\times \cdots \times \tau_K\times n_K-\tau_K+1}$ is a $2K$ order tensor for some $\tau = (\tau_1,\ldots,\tau_K),$ which is the duplication parameter of the Hankel transform (see \cite{missing_slice}).
The fixed-rank dual problem of \eqref{eqn:hankel_primal} is
\begin{equation}\label{eqn:hankel_outer_manifold_problem}
\underset{U\in S^{n_1-\tau_1+1}_{r_1}\times\cdots\times S^{n_K-\tau_K+1}_{r_K}}{\min}\;g(U),
\end{equation}
where $U = (U_1,\ldots,U_K),$ and 
\begin{equation}
\begin{aligned}
\label{eqn:hankel_inner_strong_convex_problem_constrained}
g(U) = \underset{\Z\in\mathcal{C},\s}{\max} \innerproduct{\Z}{\Y_\Omega} - \dfrac{1}{4C}\|\Z\|^2 -\sum_{k=1}^K \frac{\lambda_k}{2} \Norm{U_k^TS_k}^2, \quad \text{subject to} \quad\h^*(\s) = \Z.
\end{aligned}
\end{equation}

The problem \eqref{eqn:hankel_inner_strong_convex_problem_constrained} can be solved using linear conjugate gradient algorithm over $\Z$ and $\s,$ and the equality constraint is ensured at each step by performing a projection step. The cost $g(U)$, the gradient $\nabla g(U),$ and the directional derivative $D\nabla g(U)[V]$ can be easily computed following Lemma \ref{lemma:grad_hessian} and Remark \ref{remark:Zsdot} above. We employ Riemannian trust-region algorithm to solve \eqref{eqn:hankel_outer_manifold_problem}. The per-iteration complexity of the proposed algorithm is $
O\bigg(TI|\Omega|\sum_{k=1}^K{r_k}+\sum_{k=1}^K n_k r_k^2+\sum_{k=1}^K r_k^3\bigg)$, where $T$ is the iterations of CG and $I|\Omega|$ is the number of non-zero entries of $\s$. 

\section{Toy Experiment}

\begin{figure}[h]
\centering		
\subfigure[Gradient norm v/s Iters]{\includegraphics[width=0.5\linewidth]{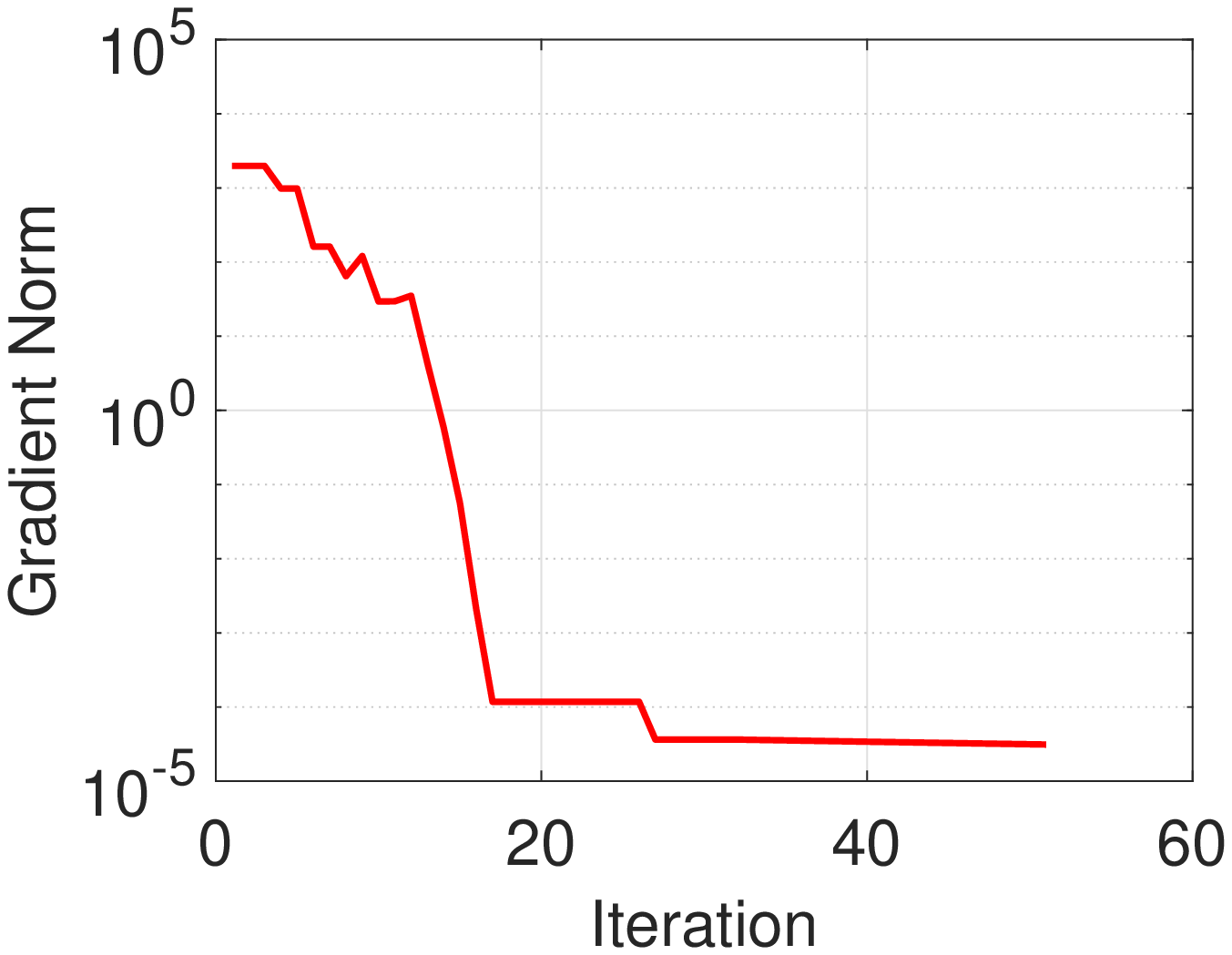}}%
\subfigure[Relative duality gap v/s Iters]{\includegraphics[width=0.5\linewidth]{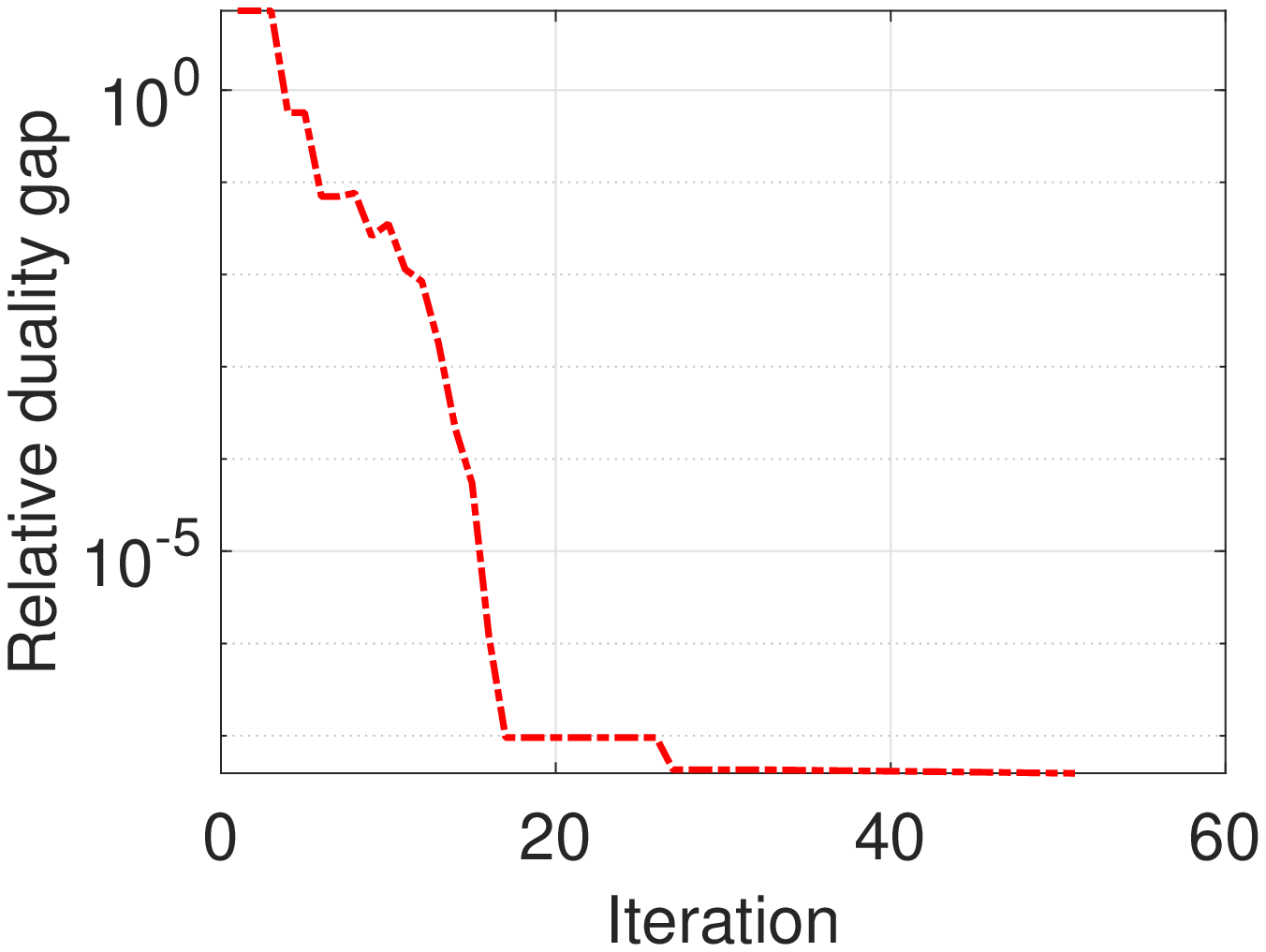}}%
\hfill
\caption{Variation of gradient norm and relative duality gap  with iterations for Hankel Tensor Completion problem.}
\label{fig:hankel_plots}
\end{figure}

We consider a toy problem of size $100\times 100\times 3$ with 10\% train data for both Nonnegative and Hankel Tensor Completion problems. The ranks $(r_1,r_2,r_3)$ are chosen as $(10,10,3)$ for  both the experiments, and the regularization constants $\lambda_k$'s are chosen according to \cite{dual}. The maximum iterations for Riemannian optimization problem was set to 200. The duplication parameter $\tau$ in the Hankel transform was set to $(10,10,1)$. The plots of the variation in gradient norm and relative duality gap with iterations for both problems is shown in Appendix \ref{sec:plots} in Figures \ref{fig:nonneg_plots} and \ref{fig:hankel_plots}. We observe that in both the cases the plots of gradient norm, and the relative duality gap decreases rapidly with iterations.


\newpage
\clearpage
\appendix

\section{Experimental plots}\label{sec:plots}

\begin{figure}[h]
\centering		
\subfigure[Gradient norm v/s Iters]{\includegraphics[width=0.5\linewidth]{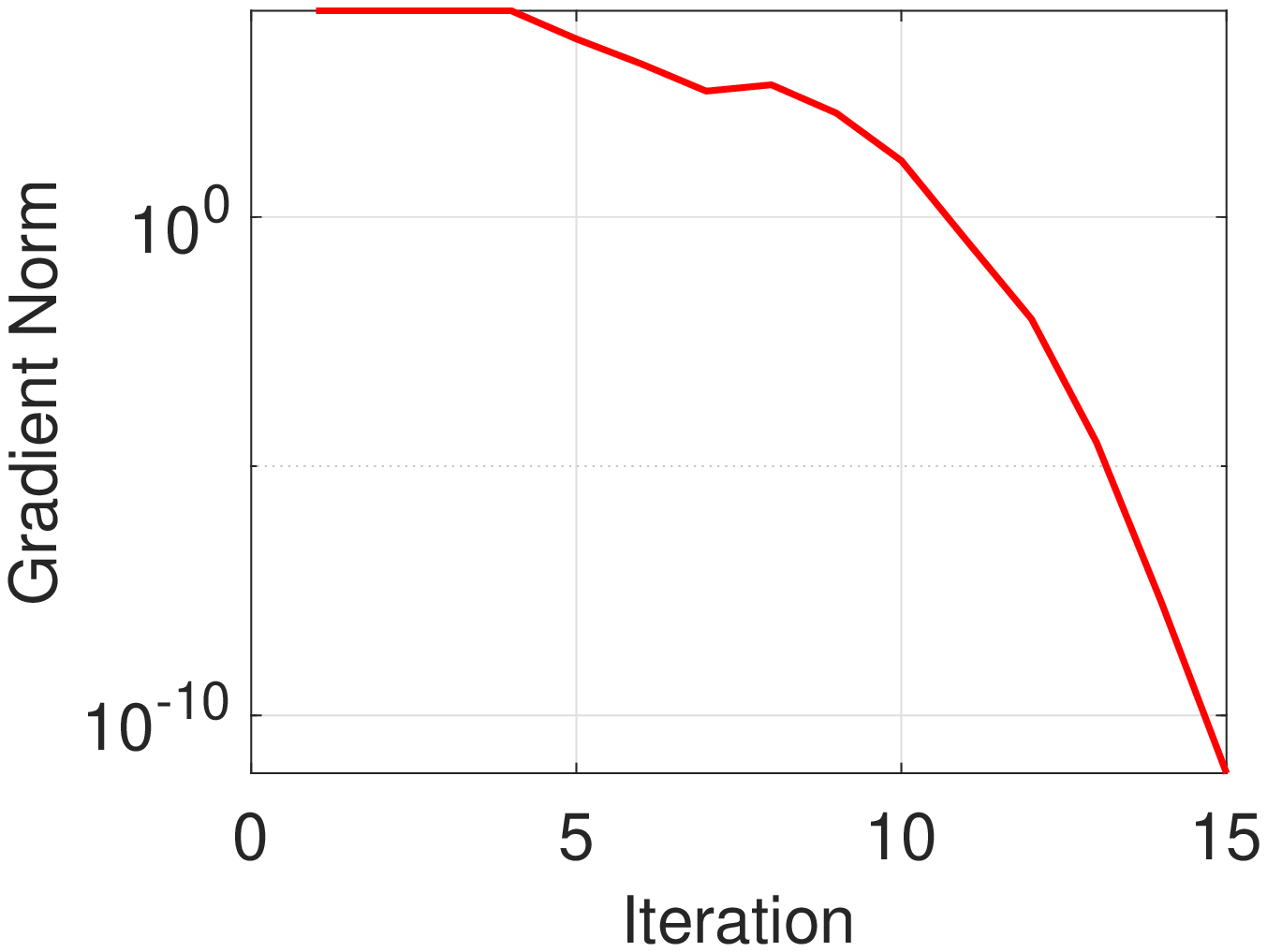}}%
\subfigure[Relative duality gap v/s Iters]{\includegraphics[width=0.5\linewidth]{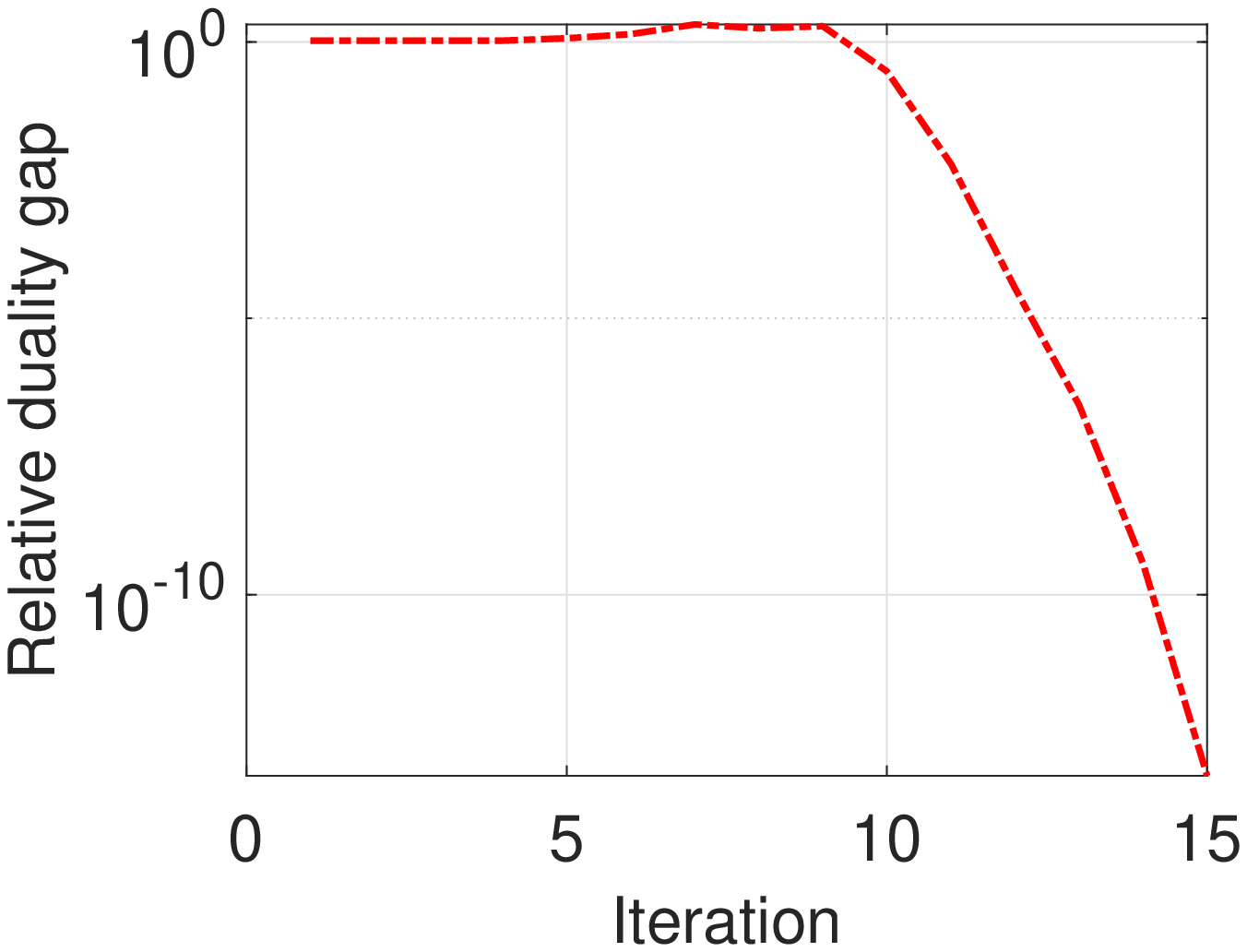}}%
\hfill
\caption{Variation of gradient norm and relative duality gap  with iterations for Nonnegative Tensor Completion problem.}
\label{fig:nonneg_plots}
\end{figure}

\section{Proof of Dual Formulation (Proof of Theorem \ref{thm:dual_form}).}

\begin{proof}\label{sec:dual_proof}
Consider the following subproblem of the above problem
\begin{equation}
\begin{aligned}
\underset{\substack{\W^{(k)},\\ k \in \{1,\ldots,K\}}}{\text{min}}& &&C\Norm{\bigg(\sum_{k=1}^K \W^{(k)}\bigg)_\Omega-\Y_\Omega}^2 + \sum_{k=1}^K \frac{1}{2\lambda_k} \innerproduct{\Theta_k^\dagger W^{(k)}_k}{W^{(k)}_k}. \\
\text{subject to}&\hspace{0.5cm} &&A(\W) \ge 0.
\end{aligned}
\end{equation}
We introduce auxiliary variables $U_k$ with the associated constraints $U_k = W_k^{(k)}$. Now, we construct the Lagrangian of the problem
\begin{equation}
\begin{aligned}\label{eqn:lagrangian}
    \mathcal{L}(\W^{(1)},\ldots,\W^{(K)},U_1,&\ldots,U_K, \Lambda_1,\ldots, \Lambda_K, s) = C\Norm{\bigg(\sum_{k=1}^K \W^{(k)}\bigg)_\Omega-\Y_\Omega}^2 + \\
    &\sum_{k=1}^K \frac{1}{2\lambda_k} \innerproduct{\Theta_k^\dagger U_k}{ U_k} + \sum_{k=1}^K \innerproduct{\Lambda_k}{W^{(k)}_k - U_k} - \innerproduct{s}{A(\W)},
\end{aligned}
\end{equation} where $\Lambda_1,\ldots, \Lambda_K,s$ are the Lagrange multipliers.
The dual function of the above will be given by 
\begin{alignat}{3}
q(\Theta_1,\ldots,\Theta_K,\Lambda_1,\ldots, \Lambda_K,s) =\underset{\substack{\W^{(k)}, U_k,\\ k \in \{1,\ldots,K\}}}{\text{min}} \mathcal{L}(\W^{(1)},\ldots,\W^{(K)},U_1,\ldots,U_K, \Lambda_1, \ldots, \Lambda_K, s).\nonumber
\end{alignat}
Taking the derivative of the Lagrangian with respect to $\W^{(k)}$ and equating to 0, we get 
\begin{align}
    \frac{\partial \mathcal{L}}{\partial \W^{(k)}} = 0
    &\implies 2C\bigg[\bigg(\sum_{k=1}^K \W^{(k)}\bigg)_\Omega-\Y_\Omega\bigg] + \fold{k} (\Lambda_k) - A^*(s) = 0 \nonumber\\
    &\implies \fold{k} (\Lambda_k) = 2C\bigg[\Y_\Omega-\bigg(\sum_{k=1}^K \W^{(k)}\bigg)_\Omega\bigg] + A^{*}(s)\label{eqn:lambda_k},
\end{align} 
where $A^*$ is the adjoint of $A$.
The right-hand side of \eqref{eqn:lambda_k} is independent of $k$, so $\fold{k}(\Lambda_k)$ is also independent of $k$. Then we have,
\begin{equation}\label{eqn:lambda_Z}
\fold{k}(\Lambda_k) = \Z + A^*(s),\text{ where }\Z = 2C\bigg[\Y_\Omega-\bigg(\sum_{k=1}^K \W^{(k)}\bigg)_\Omega\bigg].    
\end{equation}
Next, we minimize w.r.t. $U_k$, to get 
\begin{equation}\label{eqn:U_k}
    U_k = \lambda_k\Theta_k \Lambda_k.
\end{equation}
Now we compute \eqref{eqn:lagrangian} term by term by using \eqref{eqn:lambda_Z} and \eqref{eqn:U_k} 
\begin{equation}\label{eqn:term1}
C\Norm{\bigg(\sum_{k=1}^K \W^{(k)}\bigg)_\Omega-\Y_\Omega}^2 = C\Big(\dfrac{1}{4C^2}\|\Z\|^2\Big) = \dfrac{1}{4C}\|\Z\|^2,
\end{equation}
\begin{align}\label{eqn:term2}
\sum_{k=1}^K \frac{1}{2\lambda_k} &\innerproduct{\Theta_k^\dagger U_k}{U_k}-\innerproduct{\Lambda_k}{U_k} = \frac{1}{2\lambda_k} \innerproduct{\Theta_k^\dagger U_k-2\lambda_k\Lambda_k}{U_k} \nonumber\\
&= \sum_{k=1}^K \frac{1}{2\lambda_k} \innerproduct{\Theta_k^\dagger \lambda_k\Theta_k[Z_k + (A^*(s))_k]-2\lambda_k[Z_k + (A^*(s))_k]}{\lambda_k\Theta_k[Z_k + (A^*(s))_k]} \nonumber\\
&= -\sum_{k=1}^K \frac{\lambda_k}{2} \innerproduct{Z_k + (A^*(s))_k}{\Theta_k[Z_k + (A^*(s))_k]},
\end{align}
\begin{align}\label{eqn:term3}
\sum_{k=1}^K \innerproduct{\Lambda_k}{W^{(k)}_k} - \innerproduct{A^*(s)}{\W} &= \sum_{k=1}^K \innerproduct{\fold{k}(\Lambda_k)}{\W^{(k)}} - \innerproduct{A^*(s)}{\W} \nonumber\\
&= \innerproduct{\Z+A^*(s)}{\sum_{k=1}^K \W^{(k)}} - \innerproduct{A^*(s)}{\W}\nonumber\\
&= \innerproduct{\Z+A^*(s)}{\W} - \innerproduct{A^*(s)}{\W} = \innerproduct{\Z}{\W} \overset{(*)}{=} \innerproduct{\Z}{\W_\Omega}\nonumber\\
&= \bigg\langle \Z,\bigg(\sum_{k=1}^K \W^{(k)}\bigg)_\Omega\bigg\rangle \nonumber \\
&= \innerproduct{\Z}{\Y_\Omega-(1/2C)\Z} = \innerproduct{\Z}{\Y_\Omega} - \dfrac{1}{2C}\|\Z\|^2,
\end{align}
where $(*)$ is obtained by noting that $\Z = \Z_\Omega$ by definition in \eqref{eqn:lambda_Z}. Summing \eqref{eqn:term1}, \eqref{eqn:term2}, \eqref{eqn:term3}, we obtain the expression for the dual function as
\begin{equation}
q(\Theta_1,\ldots,\Theta_K,s) = \innerproduct{\Z}{\Y_\Omega} - \dfrac{1}{4C}\|\Z\|^2 -\sum_{k=1}^K \frac{\lambda_k}{2} \innerproduct{Z_k + (A^*(s))_k}{\Theta_k[Z_k + (A^*(s))_k]}.\nonumber
\end{equation}
This gives the minimax problem \eqref{eqn:minimax_problem}. From \eqref{eqn:lambda_Z} and \eqref{eqn:U_k}, we can deduce the said relation between optimal points of primal and minimax problems.
\end{proof}

\section{Proof of Duality Gap (Proof of Theorem \ref{theorem:duality_gap}).}\label{sec:duality_gap_proof} 

\begin{proof}
Following \cite{dual}, we rewrite problem \eqref{eqn:minimax_problem} in the form 

\begin{alignat*}{3}
\underset{\Theta \in P^{n_1} \times \cdots \times P^{n_K}}{\text{min}}& &&G_1(\Theta), \nonumber
\end{alignat*} where 
\begin{alignat*}{3}
G_1(\Theta) = \underset{\Z\in\mathcal{C}, s\in\R^n}{\text{max}}&\; && \innerproduct{\Z}{\Y_\Omega} - \dfrac{1}{4C}\|\Z\|^2 -\sum_{k=1}^K \frac{\lambda_k}{2} \innerproduct{Z_k + (A^*(s))_k}{\Theta_k[Z_k + (A^*(s))_k]}.
\end{alignat*}
Now, using the min-max interchange, the max-min problem equivalent to \ref{eqn:minimax_problem} can be written as 
\begin{alignat*}{3}
\underset{\Z\in\mathcal{C}, s\in\R^n}{\text{max}}&\; && G_2(\Z,s),
\end{alignat*} 
where 
\begin{alignat*}{3}
G_2(\Z,s) &= \underset{\Theta \in P^{n_1} \times \cdots \times P^{n_K}}{\text{min}} \innerproduct{\Z}{\Y_\Omega} - \dfrac{1}{4C}\|\Z\|^2 -\sum_{k=1}^K \frac{\lambda_k}{2} \innerproduct{Z_k + (A^*(s))_k}{\Theta_k[Z_k + (A^*(s))_k]} \\
&=  \innerproduct{\Z}{\Y_\Omega} - \dfrac{1}{4C}\|\Z\|^2 -\sum_{k=1}^K  \frac{\lambda_k}{2} \underset{\Theta_k \in P^{n_k}}{\text{max}} \innerproduct{Z_k + (A^*(s))_k}{\Theta_k[Z_k + (A^*(s))_k]}\\
&=  \innerproduct{\Z}{\Y_\Omega} - \dfrac{1}{4C}\|\Z\|^2 -\sum_{k=1}^K  \frac{\lambda_k}{2} B_k(\Z,s),
\end{alignat*} 
where 
\begin{alignat*}{3}
B_k(\Z,s) = \underset{\Theta_k \in P^{n_k}}{\text{max}}&\; && \innerproduct{Z_k + (A^*(s))_k}{\Theta_k[Z_k + (A^*(s))_k]}.
\end{alignat*}
We know that the value of $B_k(\Z,s)$ is equal to the largest eigenvalue of $(Z_k + (A^*(s))_k)(Z_k + (A^*(s))_k)^T$. Equivalently, if $\sigma_k$ is the largest singular value of $Z_k + (A^*(s))_k$, then $B_k(\Z,s) = \sigma_k^2$. Now, the duality gap will be given by 
\begin{equation*}    
\Delta = G_1(\hat{\Theta}) - G_2(\hat{\Z},\hat{s}) = \sum_{k=1}^{K} \frac{\lambda_k}{2} \left( \sigma_k^2 -  \|\hat{U}_k^T(\hat{Z}_k+\hat{A}_k)\|^2 \right),
\end{equation*} where $\hat{U}, \hat{Z},$ and $\hat{s}$ were defined in Theorem \ref{theorem:duality_gap}.
\end{proof}

\end{document}